 \documentclass[letterpaper, 10pt, conference]{ieeeconf}      

\IEEEoverridecommandlockouts                              

\usepackage{mathptmx} 
\usepackage{amsmath} 
\usepackage{amssymb}  

\usepackage[pdftex]{graphicx}
\usepackage{amsmath}
\usepackage{array}
\usepackage{comment}
\usepackage[utf8]{inputenc}
\usepackage{hyperref}
\usepackage{lineno}
\usepackage{microtype}
\usepackage{graphicx}
\usepackage{color}
\usepackage{amssymb}

\usepackage[infoshow,debugshow]{tabularx}
\usepackage{booktabs}
\usepackage{epsfig}
\usepackage{times}
\usepackage{amsmath}
\usepackage{float}
\usepackage{algorithm}

\usepackage{multirow}
\usepackage{lipsum}  
\usepackage{xcolor}

\usepackage{tikz}

\usepackage{booktabs}
\usepackage{multirow}
\usepackage{colortbl}
\usepackage{pgfplotstable}

\definecolor{Gray}{gray}{0.9}
\definecolor{battleshipgrey}{rgb}{0.52, 0.52, 0.51}

\usepackage{tikz}

\usepackage{pgfplots}
\usepackage{tikz}
\usepackage{graphicx}
\usepackage{pgfplots}
\usepackage{xcolor-material}
\usepackage{adjustbox}
\usepackage{graphicx}
\usepackage{pgfplots}
\pgfplotsset{compat=1.10}
\usepgfplotslibrary{fillbetween}

\usepackage[T1]{fontenc}
\usepackage{lmodern}

\usepackage{url}
\DeclareUrlCommand\function{\urlstyle{sf}}
\usepackage{xcolor}
\usepackage{tikz}
\usepackage{booktabs}
\usepackage{array}
\usepackage{multirow}
\usepackage{siunitx}

\definecolor{chart Offline}{gray}{.5}
\definecolor{chart Online}{RGB}{227,227,138}
\definecolor{chart Ok}{RGB}{248,172,37}
\definecolor{chart Ideal}{RGB}{1,151,0}
\definecolor{chart Over}{RGB}{0,125,234}

\newdimen\tempdim

\newcommand*{\ChartLegend}[1]{%
  \ifdim\lastkern=1sp %
    \hspace{1em}%
  \fi
  \ChartBox{0.75em}{#1}%
  \,#1%
  \kern-1sp\kern1sp\ignorespaces
}
\newcommand*{\ChartBox}[2]{%
  \begingroup
    \settoheight{\tempdim}{L}%
    \edef\tempheight{\the\tempdim}%
    \settodepth{\tempdim}{g}%
    \edef\tempdepth{\the\tempdim}%
    \tikz[
      baseline=0pt,
      inner sep=0pt,
    ]
    \node[
      fill={chart #2},
      draw,
      rounded corners=1pt,
      anchor=base,
    ]{%
      \vphantom{g\"A}%
      \pgfmathsetlength{\tempdim}{(#1)}%
      \pgfmathsetmacro{\len}{#1*0.04}
      \kern\tempdim\relax
    };%
  \endgroup
}

\usepackage{pgfplots}
\usepackage{datatool}
\usetikzlibrary{matrix}
\usetikzlibrary{patterns}
\usepgfplotslibrary{groupplots}
\pgfplotsset{compat=newest}

\usepackage[noend]{algpseudocode}
\algdef{SE}[DOWHILE]{Do}{doWhile}{\algorithmicdo}[1]{\algorithmicwhile\ #1}%
\algnewcommand\algorithmicswitch{\textbf{switch}}
\algnewcommand\algorithmiccase{\textbf{case}}

\algnewcommand\algorithmicassert{\texttt{assert}}
\algnewcommand\Assert[1]{\State \algorithmicassert(#1)}%
\algdef{SE}[SWITCH]{Switch}{EndSwitch}[1]{\algorithmicswitch\ #1\ \algorithmicdo}{\algorithmicend\ \algorithmicswitch}%
\algdef{SE}[CASE]{Case}{EndCase}[1]{\algorithmiccase\ #1}{\algorithmicend\ \algorithmiccase}%
\algtext*{EndSwitch}%
\algtext*{EndCase}%

\newcommand{\rev}[1]{{#1}}
\graphicspath{{img/}}

\begin{document}
\title{Bringing Online Egocentric Action Recognition\\ into the wild}

\author{Gabriele Goletto$^{*,1}$, Mirco Planamente$^{*,1,2,3}$, Barbara Caputo$^{1,3}$, and Giuseppe Averta$^{1}$
\thanks{This work was supported by the Italian Ministry of University and Research (DM1061), the IIT HPC infrastructure for the availability of high performance computing (Franklin) and CINECA award under the ISCRA initiative. This study was carried out within the FAIR - Future Artificial Intelligence Research and received funding from the European Union Next-GenerationEU (PIANO NAZIONALE DI RIPRESA E RESILIENZA (PNRR) – MISSIONE 4 COMPONENTE 2, INVESTIMENTO 1.3 – D.D. 1555 11/10/2022, PE00000013). This manuscript reflects only the authors’ views and opinions, neither the European Union nor the European Commission can be considered responsible for them.}      
\thanks{$^{*}$The authors equally contributed to this work.}
\thanks{$^{1}$  Dipartimento di Automatica e Informatica, Politecnico di Torino, Corso Duca degli Abruzzi, 24, 10124 Torino, Italy
{\tt\small name.surname@polito.it}}
\thanks{$^{2}$ Italian Institute of Technology, Genova, Italy}
\thanks{$^{3}$ Consortium Cini, Italy}
%
}


\maketitle

\definecolor{greeno}{rgb}{0,0.502,0}
\definecolor{orangee}{RGB}{236, 173, 64}
\definecolor{bluee}{RGB}{118, 159, 182}
\definecolor{grayy}{HTML}{BAD7F2}
\definecolor{pal1}{HTML}{433a3f}
\definecolor{pal2}{HTML}{6ba368}
\definecolor{pal3}{HTML}{fce762}

\begin{abstract}

To enable a safe and effective human-robot cooperation, it is crucial to develop models for the identification of human activities. Egocentric vision seems to be a viable solution to solve this problem, and therefore many works provide deep learning solutions to infer human actions from first person videos. However, although very promising, most of these do not consider the major challenges that comes with a realistic deployment, such as the portability of the model, the need for real-time inference, and the robustness with respect to the novel domains (i.e., new spaces, users, tasks). With this paper, we set the boundaries that egocentric vision models should consider for realistic applications, defining a novel setting of egocentric action recognition in the wild, which encourages researchers to develop novel, applications-aware solutions. We also present a new model-agnostic technique that enables the rapid repurposing of existing architectures in this new context, demonstrating the feasibility to deploy a model on a tiny device (Jetson Nano) and to perform the task directly on the edge with very low energy consumption (2.4W on average at 50 fps). \rev{The code is publicly available
at: \small{\url{https://github.com/EgocentricVision/EgoWild}}}.
\end{abstract}


\IEEEpeerreviewmaketitle
\section{Introduction}
Current robotics research demonstrated an increasing interest in the development of technologies to support the physical interaction between humans and machines, ranging from the planning and control \cite{ajoudani2018progress}, up to their social impact \cite{henschel2020social}. However, the deployment of this technology in the real world 
requires an extension of the human intention retrieval capabilities of robots, from a mere pose estimation and forecast, to an high level description of the action executed. 
To reach this goal, a very promising solution relies on the use of egocentric vision, in which the human activity is recorded by wearable cameras placed on the head of the user \cite{rodin2021predicting}. 
This setting comes with the benefit that source data are characterised by a rich multi-modal information, thanks to the proximity of audio/video sensors to the action scene, and by an intrinsic embedding of an attention mechanisms that stems from the human gaze direction itself.


\begin{figure}
    \centering
    \includegraphics[width=1\linewidth]{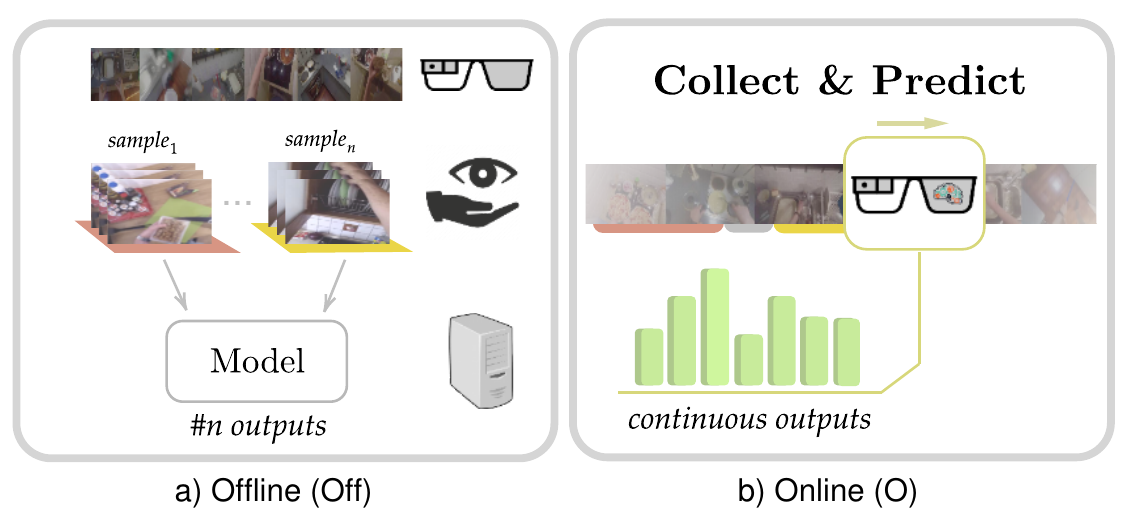}
    \begin{flushright}
    \pgfplotsset{compat=1.9}
\pgfplotsset{mystyle/.style={%
scatter, scatter/use mapped color={draw opacity=0,fill=mapped color}, only marks, point meta=explicit symbolic,fill opacity=0.8, text opacity=1,mark=*, scatter/classes={
            a={draw=grayy, fill=grayy, mark size=0.5*3.7, label=2},%
            b={draw=MaterialOrange800, fill=MaterialOrange800,mark size=2.4},%
            c={draw=grayy, fill=grayy,mark size=0.38*24.9},%
            d={draw=MaterialOrange800, fill=MaterialOrange800,mark size=0.5*11.5},%
            e={draw=grayy,fill=grayy, mark size=0.3*53.7},%
            f={draw=MaterialOrange800, fill=MaterialOrange800,mark size=0.2*131.8}%
          }}}

\pgfplotsset{
  standard/.style={
        axis x line=middle,
        axis y line=middle,
        enlarge x limits=0,
        enlarge y limits=0,
        every axis x label/.style={at={(ticklabel* cs:0.9)},anchor=north west},
        every axis y label/.style={at={(ticklabel* cs:1.15)},anchor=north, xshift=-4ex}
    }
  }

\begin{tikzpicture}
\footnotesize
      \begin{axis}[
          width=2.5in,
          height=1.3in,
          xtick=data,
          symbolic x coords = {0,A,1,B,2,C,3,4},
          ylabel=FPS,
          ylabel style={left},
          xlabel=Device Size,
          ytick={3,30, 300},
          xmin=0,
          xmax=4,
          ymax=700,
          ymin=1,
          xtick distance=0.5,
          axis y line*=left,
          axis x line*=bottom,
          axis lines = middle,
          ymode=log,
          log ticks with fixed point,
          scale only axis,
          xticklabels={RTX 2080Ti,MX350,Jetson Nano},
          standard
       ]
      \addplot[mystyle] table [x={x},y={Yval},meta index=4] {
x  Algorithm Val Yval Label
A  I3D  92.7 110 e
B  I3D  24.9 15 c
C  I3D  3.7 3 a
};      
\node (A) at (570,1.2) {{\textcolor{black}{3.7W}}};
\node (C) at (310,1.6) {{\textcolor{black}{24.9W}}};

\addplot[pal1,line legend, dashed, sharp plot,nodes near coords={},
    update limits=false]
    coordinates {(0,30) (4,30)};
\addplot[pal1,line legend, dashed, sharp plot,nodes near coords={},
    update limits=false,shorten <=-20mm, shorten >=-4mm] 
    coordinates {(2,500) (2,2)};
\addplot[pal1,line legend, dashed, sharp plot,nodes near coords={},
    update limits=false,shorten <=-20mm, shorten >=-4mm] 
    coordinates {(1,500) (1,2)};

\addplot[pattern=north east lines, pattern color=MaterialOrange200, draw=none, opacity=.5] 
    coordinates {(0,30) (0,700) (1,700) (1,30)};
    
\addplot[fill=pal2, draw=none, fill opacity=.1,
   update limits=false] 
    coordinates {(1,30) (1,700) (2,700) (2,30)};


\addplot[pattern=north east lines, pattern color=MaterialOrange200, draw=none, opacity=.5] 
    coordinates {(2,30) (4,30) (4,1) (2,1)};
\addplot[pattern=north east lines, pattern color=MaterialOrange200, draw=none, opacity=.5] 
    coordinates {(1,30) (2,30) (2,1) (1,1)};
\addplot[fill=pal2, draw=none, fill opacity=.1,
   update limits=false] coordinates {(2,700) (4,700) (4,30) (2,30)};

\addplot[pattern=north east lines, pattern color=MaterialOrange200, draw=none, opacity=.5] coordinates {(0,30) (1,30) (1,1) (0,1)};

\end{axis}
\node[above, font=\sffamily] at (0.9,3.3){Workstation};
\node[above, font=\sffamily] at (2.8,3.3){Portable};
\node[above, font=\sffamily] at (5,3.3){Edge};
\node[above, font=\sffamily] at (6.8,2.3){Online};
\node[above, font=\sffamily] at (6.8,0.7){Offline};
\end{tikzpicture}
\begin{tikzpicture}
\footnotesize
\node[overlay] (E) at (-6.6,2.85) {{\textcolor{black}{53.7W}}};
\end{tikzpicture}
    \end{flushright}
     


  
%
%

\caption[Caption for LOF]{\textbf{Top:} Comparison between offline (a) and online (b) \rev{inference} protocol for first person action recognition (FPAR). \textbf{Bottom}: Frames per Second (FPS) processed with the I3D model \cite{carreira2017quo} on different devices. 
The \protect\tikz \protect\node [rectangle, pattern=north east lines, pattern color=MaterialOrange200, draw=MaterialOrange200, opacity=1] at (2.5,-2) {}; areas show traditional action recognition models' difficulty to run online \rev{inference} on edge devices, either due to latency or hardware constraints. 
Our goal is to promote research toward models that can work in the \protect\tikz \protect\node [rectangle,draw=none, fill=pal2, opacity=.5] at (2.5,-2) {}; area, allowing egocentric models to run online \rev{inference} and on tiny devices.
}
\label{fig:teaser}
\vspace{-0.6cm}
\end{figure}

\renewcommand{\thempfootnote}{\fnsymbol{footnote}}
\renewcommand{\thefootnote}{\fnsymbol{footnote}}

Although many works in the literature have provided solutions to infer knowledge about human activity from egocentric data (a.k.a. First Person Action Recognition, FPAR), this is frequently achieved through very large neural architectures without regard to their computational demand (see Fig. \ref{fig:teaser}, bottom part). As a consequence, although very accurate, most of the models presented in literature 
are not suitable for realistic usecases, where real-time inference (Fig. \ref{fig:teaser}, online scenario) should be performed on board of low-power hardware to enable wearability, avoid data transfer and preserve privacy.
The goal of this paper is to encourage a new line of research based on realistic egocentric vision use cases. We propose a new FPAR benchmark with real-world constraints, which consists of altering current action recognition protocol to follow a set of realistic limitations that we add progressively (model size, cross domains, online, and untrimmed). 

In addition, we propose a model-agnostic technique to enable a fast re-purposing of existing architectures in this new context. Our approach consists of two components: an anomaly detection-based solution for action boundary localization, followed by a two-fold aggregator strategy. The first solution is based on the assumption that \textit{if I can recognize an action, I can also localize it}. Considering that traditional training of the action recognition framework is done with trimmed data containing single actions, the embedding that arises from multiple actions will be very different from the standard one, and as a consequence, the network will be able to detect it as an anomaly. The second solution is introduced to cope with the large proportion of overlapping segments in fine-grained action recognition that make it harder to localize concurrent actions.

To summarize, \rev{this paper contributes with:}

\begin{itemize}
    \item \rev{the definition of a new setting of FPAR in the wild, which encourages researchers to develop applications-aware solutions;}

    \item \rev{a benchmark of popular action recognition models for real-world application in FPAR;}

    \item \rev{a method to enable the use of existing features extractors to  achieve efficient yet accurate action recognition under constraints, exploiting an anomaly detection strategy to localize the boundary of the actions and a two-fold aggregator solution to deal with concurrent actions in a continuous stream;}

    \item \rev{an analysis of performance on an edge device, opening interesting perspectives for on-board intelligence.}
   
\end{itemize}

\section{Related Works}
\label{RW}

\begin{figure}[t]
\vspace{2mm}
 \includegraphics[width=0.49\linewidth]{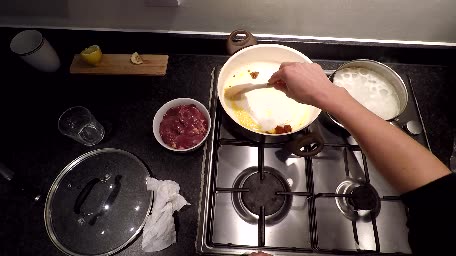}
 \includegraphics[width=0.49\linewidth]{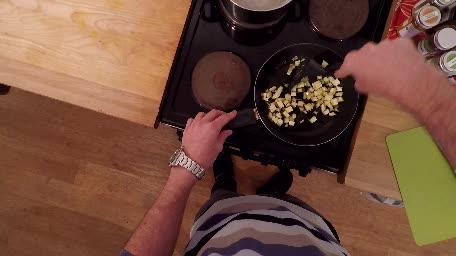}
\caption{Examples of two frames taken from different videos sources with the same label (``mix'').}
\vspace{-0.5cm}
\label{img:unseen}
\end{figure}

\par \textbf{First Person Action Recognition (FPAR).}
The main architectures utilized in this context are generally inherited from third-person literature and may be classified into two broad categories: 2D convolution -based \cite{wang2016temporal, lin2019tsm, zhou2018temporal, cartas2019seeing} and 3D convolution -based \cite{carreira2017quo,singh2016first, tran2015learning, feichtenhofer2019slowfast, Munro_2020_CVPR}. The first group is generally complemented with other modules such as LSTM or its variations \cite{sudhakaran2019lsta, furnari2020rolling, planamente2021self}, Temporal Shift Module (TSM) - a parameter-free channel-wise temporal shift operator presented in \cite{lin2019tsm}, or the Temporal Relation Network module (TRN) \cite{zhou2018temporal}.
The use of 3D convolutions was proposed as an alternative in~\cite{carreira2017quo, tran2015learning} to learn spatial and temporal relations simultaneously, even if  they often introduce more parameters, requiring pre-training on large-scale video datasets~\cite{carreira2017quo}.
The complex nature of egocentric videos raises a variety of challenges, such as ego-motion~\cite{li2015delving}, partially visible or occluded objects, and environmental bias~\cite{munro2020multi, planamente2021domain}, \rev{\cite{planamente2023toward}}, which limit the performance of traditional approaches when used in FPAR 
~\cite{damen2018scaling, damen2021rescaling}. 
Those challenges attract the community's interest and motivate the design of novel and more complex architectures, often based on multi-stream approaches such as \cite{sudhakaran2019lsta, wang2021interactive, kazakos2021little, furnari2020rolling}.


\par \textbf{Action segmentation and detection.}
Action segmentation \cite{huang2016connectionist, wang2020boundary,li2021temporal,khan2022timestamp}, \rev{\cite{aakur2019perceptual, mounir2022spatio}}  and detection \cite{shou2016temporal, shou2017cdc, piergiovanni2019temporal} can be intended as the extension of action recognition to the more complex scenario of untrimmed videos, where the task is to assign an action label to each frame, identifying non-overlapping (for segmentation) and overlapping (for detection) action segments.
Most of these tasks require large and offline models, especially for the EPIC-KITCHENS challenge solutions\footnote{https://epic-kitchens.github.io/Reports/EPIC-KITCHENS-Challenges-2021-Report.pdf}, in which the network uses the entire video as input to infer the action. This makes state-of-the-art models unsuitable for our purpose, where on-line processing is fundamental. Recently, \cite{du2022fast} developed a novel unsupervised methodology for event boundary localization (detection) that outperforms current approaches while increasing inference time significantly, making it perfect for edge devices.

\section{Bringing FPAR in the wild}

To really enable the deployment of egocentric vision models, it is fundamental to consider a variety of constraints in terms of energetic, memory and temporal budget. The first (and foremost) of these is the amount of resources required to perform the task, namely the memory size to store model parameters and input data, and the number of operations (e.g. MACs) required to perform inference. The first is a constraint imposed by the flash memory of the device, while the second is related to the micro-controller velocity in inference, and to the frame-rate required by the task. 

The input specification is another important feature to consider for real-world applications. In this regard, the goal is to find a good trade-off between: i) the amount of information needed as input to properly encode the temporal information; ii) the corresponding memory increase for storing input data on the device; and iii) the critical fact that, unlike the spatial dimension, the temporal dimension 
is presented as a continuous stream, which prevents an efficient sub-sampling and requires online processing. Another important aspect to consider when posing real-time constrains is that, in the context of egocentric vision, many techniques attain notable results only by leveraging non-real-time secondary modalities such as the optical flow. Although this modality is highly successful, it has a high computational cost \cite{crasto2019mars, plizzari2022e2}, which prevents its use in real-time applications, and increases the size of the model.

\rev{It is also worth reporting that, because the sensor is worn by the user - usually at the head level - it records data with a high degree of variation produced by rapid changes in environment, perspective, and illumination as in Fig \ref{img:unseen}. 
Input variability can cause a difference in the distribution of data between the training and testing phases. This results in a problem known as \textit{environmental bias} or \textit{domain shift} that can negatively impact the performance of the model.}
Studying the network capability to generalize across domains provides clues on how the model will perform in a real scenario (where domain shifts are present).


The last point of interest for a real deployment of egocentric technologies in the wild \rev{lies in the intrinsic untrimmed nature of input data.} Indeed, the vast majority of works of action recognition assume that the input clips are  ``trimmed''  around the action of interest, which clearly represent an invasive form of supervision not available in realistic settings. Therefore, we argue that, despite recent progress in the area, trimmed action recognition has limited relevance in real-world scenarios, while continuous video flows with no previous knowledge on action location in time are the primary input source to be considered. 

Model size, online recognition, robustness across domains, and untrimmed data source represent the constraints that realistic usecases pose, and - to the best of our knowledge - no work in the literature investigates general solutions appropriate for this setting. In this paper, beyond proposing a new line of research, we investigate a solution to bring existing  FPAR models to perform online action recognition without introducing further training, thereby promoting the repurposing of existing models.

\subsection{Benchmarking FPAR with real-world constraints.} 
\par
%
As a first step, we tackle the model footprint issue and assess the impact of the model reduction by comparing it with popular action recognition networks, testing their generalization capabilities in seen and unseen settings (domain shift). Then, following an increasing complexity order, all real-world restrictions are added sequentially (Streaming, Online, and Untrimmed).


\begin{figure}[t]
    \centering
    \includegraphics[width=0.9\linewidth]{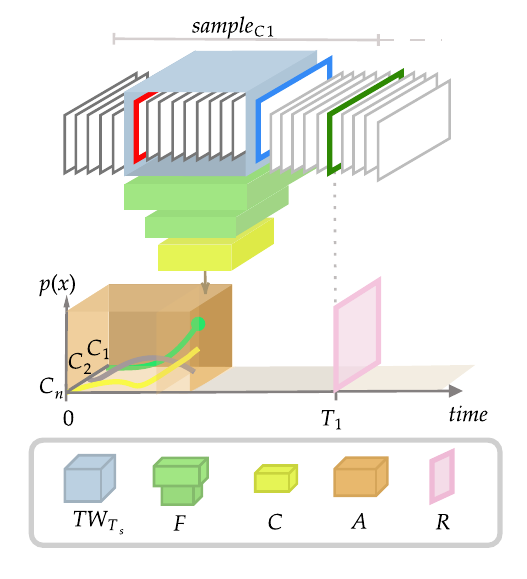}
    \vspace{-0.6cm}
    \caption{
    Illustration of the Streaming \rev{inference} scenario. $TW_{T_{s}}$ represents a temporal window sliding along the video with stride 1.
    At each time step, a clip of $T_{s}$ contiguous frames is fed into the network, which comprises a feature extractor $F$ and a classifier $C$ with $n$ classes ($C_{1}$, $C_{2}$, ... ,$C_{n}$). $A$ represents the aggregator that - at each step - updates the output of the network, taking into consideration the current output and the previous ones. R stands for aggregator cleaning, triggered by the sample's last frame.
    }
    \label{fig:overview}
    \vspace{-0.59cm}
\end{figure}

\textbf{Backbone. }
To assess the effects of model footprint on task accuracy, we considered several 
2D-CNN and 3D-CNN models for action recognition, which are often used in the context of egocentric vision, including I3D \cite{carreira2017quo}, TSN \cite{wang2018temporal}, TSM \cite{lin2019tsm} and TRN \cite{zhou2018temporal}. These typical action recognition architectures are compared with two families of NAS-based models which optimize model efficiency: \cite{feichtenhofer2020x3d} and \cite{ kondratyuk2021movinets}. From these, we considered for our purposes the smallest versions, named X3D-XS and MoViNet-A0 (with and without buffer) respectively. 
%
We tested the backbones on both seen and unseen data distributions (e.g. different environments). Albeit this is very often omitted, testing across domains is fundamental to assess the generalization capabilities of models and can highlight overfitting occurrence on specific data distributions. 

\textbf{Offline, Streaming, and Online. }
As anticipated before, the standard action recognition \rev{inference} protocol usually works in an offline fashion, exploiting the supervision information on the edges of the action (start and end) to take the right input to process. 
To do this, specific sampling strategies are needed to reduce the amount of input data and avoid that a sample $x$ of $T$ frames cause a model to exceed its memory budget. Most works address this issue relying on uniform sampling, i.e picking $T_s$ equidistant frames, with  $T_s < T$. This method is the preferable solution for video understanding, but suffers of two major drawbacks: i) it assumes the knowledge of samples length in advance (which is not the case for continuous streams); and ii) uniform sampling completely filters out information related to the action velocity. Other works, instead, rely on dense sampling, selecting a set of $T_{s}$ contiguous frames. In some cases, this choice penalizes the model due to the fact that its temporal receptive fields may see only a limited portion of the action. \rev{Indeed, the final prediction is usually obtained by averaging the predictions of different equidistant clips over the whole video, performing video level uniform sampling, i.e. requiring the sample's length information.}

%
%
%
The artificial limits of offline \rev{inference} approaches are alleviated in two novel cases. The first, hereinafter named \textit{Streaming \rev{inference}}, still assumes the knowledge of action boundaries but enables the processing of a continuous input stream (see Fig. \ref{fig:overview}). Intermediate outputs are continuously collected with an aggregator ($A$ in Fig. \ref{fig:overview}) which is then used to obtain a final prediction. When the action is completed (i.e. at the final frame), the aggregator is flushed to reset the model for novel predictions. 
%
%
%
\rev{Removing the supervision on action boundaries as well} (i.e. no prior knowledge on when to reset the aggregator), we introduce the \textit{Online \rev{inference}} setting, where the model is asked to identify both actions and their (rough) temporal edges. 
The complexity of this setting requires to deal with untrimmed data when actions are alternated with ``unknown'' clips. In our experiments, we studied the online \rev{inference} settings with and without ``unknown'' clips, to verify how their presence affects the final performance. 

\section{Methodology}

\subsection{
From single clips to continuous data streams}
We extended the offline \rev{inference} approach to deal with continuous streaming input by using a sliding window ($TW_{T_{s}}$) with a unitary stride that selects $T_{s}$ dense frames progressively (see Fig. \ref{fig:overview}). For each time sample, the oldest frame (red in Fig. \ref{fig:overview}) is removed and replaced by a new one (blue in Fig. \ref{fig:overview}), and a new inference is performed and accumulated with the previous ones. Then, a continuous output is obtained with an aggregator strategy (aggregator(A) in Fig \ref{fig:overview}).
MoViNet implements its aggregator by replacing 3D convolution with the (2+1)D operation and exposing a stream buffer mechanism to cache feature activations, allowing the temporal receptive field to expand without the need for recomputation. To support frame-by-frame output and exploit the buffer mechanism, it uses Causal Convolutions and Cumulative Global Average Pooling. The first one is used to make the convolutions unidirectional along the temporal dimension. The second one, instead, approximates any global average pooling involving the temporal dimension. For models lacking specific aggregator mechanisms, we implemented a continuous averaging of the corresponding temporal window’s output. Each aggregator is empty at the beginning of each sample and is resetted (R) at the end.

\subsection{Actions boundaries localization}
\label{eventboundarydetection}
\par 

As anticipated before, action recognition models are trained to classify well-separated actions taken as input. Transferring this capability to continuous video flows comes with the difficulty that the model may be asked to infer from clips that do not necessarily contain separate and complete actions. Therefore, the continual encoding of successive actions results in an overall increase in prediction uncertainty and instability in time. The anomaly detection literature \cite{chandola2009anomaly} describes this behavior as a consequence of the fact that the network processes data with a pattern that does not conform to the defined notion of \textit{normal} data learned during training. Therefore, the presence of concurrent or unknown action can be seen as an anomaly in time. Based on this consideration, we implemented a Dynamic Boundary Localization (DBL) strategy - with almost no overhead in terms of model size and latency - to localize the boundary of an action by examining the continuous stream of extracted features.

More specifically, since cross-entropy loss (de facto standard for FPAR) promotes class representations to be well separated in feature space \cite{wang2017normface}, it is possible to use a distance metric (e.g. Mean Square Error) between the features extracted to measure their variations caused by action changes. Therefore, it is possible to identify action boundaries in a continuous data flow by looking for abnormalities in feature distribution over time while treating all frames of the same action as \textit{normal}. This method could not only reveal differences between known actions but also detect the presence of ``unknown'' segments of video (e.g., background).

However, it is important to note that in case of overlapping segments (which e.g. in the EPIC-Kitchens dataset reaches up to 28.1\% of the total clip), the detection of the new class could be delayed or anticipated with respect to the current action. Since the aggregator solution can encode only one action at a time, the network's inference will favor one of the two consecutive actions.

\begin{figure}[t]
    \centering
    \includegraphics[width=0.95\linewidth]{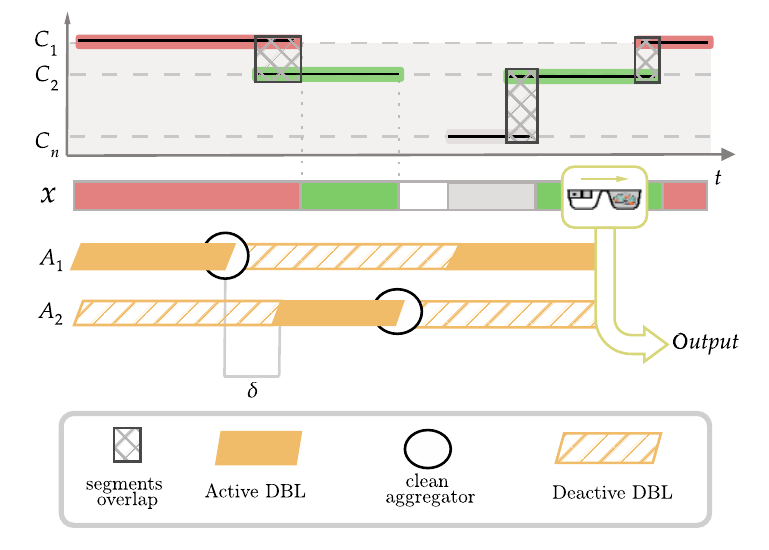}
     \vspace{-0.55cm}
    \caption{Illustration of the proposed two-fold aggregator ($A^{2}$) method. The two aggregators work asyncronsly, $\delta$ is a parameter used to guarantee the asyncroncity of the two and indicates the frame-delay of the DBL activation of one aggregator when the other one detects an anomaly.}
     \label{fig:oursolution}
    \vspace{-0.5cm}
\end{figure}

In light of the above considerations, we can state that for fine-grained action recognition, the standard aggregator may be ineffective. To solve this problem, we introduced a two-fold aggregator strategy ($A^2$). The two aggregators ($A_1$ and $A_2$ in Fig. \ref{fig:oursolution}) run asynchronously using a mixed boundary detection approach, allowing the encoding of the next action before the previous one finishes. When one aggregator detects an anomaly, it disables its DBL and activates the DBL of the second aggregator. To guarantee asynchrony in the moment of the anomaly's detection, we delay the activation of the second aggregator's DBL by an hyperparameter $\delta$.

The final output is obtained as:
\begin{equation}
\mathcal{O} = n_{1}A_{1}(x) + n_{2}A_{2}(x)
\end{equation}
where $A_{i}(x)$ is the output of the \textit{i-th} aggregator for the input $x$ and $n_{i}$ corresponds on the quantity of frame processed by the \textit{i-th} aggregator.

\section{Implementation}

\textbf{Dataset.}
\rev{In our experiments, we utilize the top three kitchens with the most labeled samples from the EPIC-Kitchens-55 dataset~\cite{damen2018scaling}. These kitchens are referred to as D1, D2, and D3. We have chosen this specific setting as it is the standard and widely used dataset for cross-domain analysis in first-person perspective~\cite{Munro_2020_CVPR}, and it also provides rich multi-modal information, including audio and event data~\cite{plizzari2022e2}, which can be beneficial for further analysis. Additionally, the difficulties in this dataset arise not only from the significant domain shift among different kitchens, but also from imbalanced class distribution both intra- and inter-domain.}

\textbf{Input.}
 Experiments with I3D~\cite{carreira2017quo} and X3D~\cite{feichtenhofer2020x3d} are conducted by sampling one random clip from the video during training and 5 equidistant clips spanning across all the video during test, as in~\cite{munro2020multi}. The number of frames composing each clip is 16. 
 For TSN~\cite{wang2018temporal}, TSM~\cite{lin2019tsm} and TRN \cite{zhou2018temporal} architectures, uniform sampling is used, consisting of 5 frames uniformly sampled along the video. During testing, 5 clips per video are adopted, following the experimental protocol proposed in ~\cite{lin2019tsm}. 
 For MoVinet \cite{kondratyuk2021movinets}, dense sampling is adopted, with 4 consecutive clips composed by 8 frames, randomly taken from the video during training as in the original work.
 All the architectures follow the standard video data augmentation as in~\cite{wang2016temporal}, the spatial input resolution has been kept consistent with the pretrained models (182 for X3D, 172 for MoViNet and 224 for the others) while the temporal resolution for all the models has been set to 30 fps.
 
\textbf{Implementation Details.}
 We adopted the original I3D network proposed in~\cite{carreira2017quo} with Inception-V1 as inflated backbone, while we chose to use X3D-XS and MoViNet-A0 to have the most efficient models from the two families. The optimizer is SGD with momentum of 0.9,  weight decay $10^{-7}$ and a starting learning rate $\eta$ of 0.01. I3D has been trained for a total of 5000 iterations, the learning rate decays by 0.1 at step 3000. Instead, MoViNet-A0 and X3D-XS have been trained for 1500 iterations without learning rate decay. For all the experiments we adopted a batch size of 128. \rev{For the two-fold aggregator implementation, we estimated the value of $\delta=20$ directly from the dataset (a subset of kitchens from \cite{damen2021rescaling} not used in this paper) by calculating the average length of action overlaps (at 30 Hz).}

\textbf{Evaluation Protocols. }
In this part, we discuss the evaluation protocol we used for our benchmark.
\\
\textbf{Seen} $\rightleftharpoons$ \textbf{Unseen}. For the seen results we train on kitchen $D_i$ and test on the same ($D_i \rightarrow D_i$), $i \in \{1,2,3\}$. We evaluate performance on unseen test by training on $D_i$ and testing on  $D_j$, with $i \neq j$ and $i, j \in \{1,2,3\}$ ($D_i \rightarrow D_j$). 
\\
\textbf{Offline, Streaming, and Online.} We  refer to \textit{offline} to indicate the standard action recognition \rev{inference} protocol, which typically uses as input a sub-sample of the input frames. We perform experiments using both uniform and dense sampling. 
The term \textit{streaming \rev{inference}} refers to experiments where the test is performed using all the frames (to simulate the continuous stream of the data that comes from a wearable device) with the supervision on the action's boundary (start and end) to properly clean the aggregator.
The \textit{online \rev{inference}} setting, instead, assumes no supervision on action limits, and to effectively deal with this scenario the model should automatically detect both action and their boundaries.

\par \textbf{Trimmed $\rightarrow$ Untrimmed.} 
Moving from the trimmed to the untrimmed scenario, the lack of mutually exclusive temporal separation from the actions found in the original dataset makes it difficult to calculate an accuracy per frame. At the same time, the precise timestamp of start and end in fine-grained action is complex and extremely subjective. For this reason, we use accuracy as a metric to validate our experiments, putting the focus on the ability to recognize the action when it happens instead of the precise localization of its boundaries. For simplicity, the performance is still computed at the end of each action, and the  ``unknown'' segments are not used in the evaluation of the performance but only in the event boundary localization part. In other words, we did not use the ``unknown'' class during the evaluation of the accuracy, but the network should be able to manage it during the clean phase of the buffer or the logits accumulation.

\setlength{\tabcolsep}{8pt}
\begin{table}[t]
\begin{center}
\vspace{3mm}
\caption{Top-1 \textit{mean} accuracy ($\%$) of different common-use architectures, over all $D_i \rightarrow D_j$ combinations on both seen and unseen test sets in \textit{offline-trimmed} setting. }
\label{tab:backbone}
\begin{tabular}{lccccc}
\toprule\noalign{\smallskip}
\multicolumn{5}{c}{\normalsize\textsc{EPIC-Kitchens 55}} \\
\noalign{\smallskip}
\cline{1-5}
\noalign{\smallskip}
Network  & Sampling & Params & Seen & Unseen \\
\noalign{\smallskip}
\toprule\noalign{\smallskip}
\multirow{1}{*}{\centering TSN} 
  & U 5x5 & 10.7M & 60.88 & 31.55 \\
 
  \noalign{\smallskip}

 \multirow{1}{*}{\centering TSN-TRN} 
  & U 5x5  & 18.3M & 63.13 & 32.42 \\

 \noalign{\smallskip}
\multirow{1}{*}{\centering TSM} 
  & U 5x5 &  24.3M & \textbf{71.48} & 35.97 \\

   \noalign{\smallskip}
\multirow{1}{*}{\centering TSM-TRN} 
  & U 5x5 & - & 69.52 & 36.05 \\
 
  \hline 
 \noalign{\smallskip}
\multirow{1}{*}{\centering I3D }
  & U 16x5 & 12.4M & 67.34 & \textbf{43.89} \\ 
 
 \noalign{\smallskip}
\multirow{1}{*}{\centering I3D }
  & D 16x5 & 12.4M & 67.08 & 42.42 \\
    \hline
 \noalign{\smallskip}
\multirow{1}{*}{\centering X3D-XS }
  & U 5x5 & 3.8M & 51.46 & 36.39 \\ 	

 \noalign{\smallskip}
\multirow{1}{*}{\centering X3D-XS }
  & D 16x5 & 3.8M & 48.45 & 32.66 \\

    \noalign{\smallskip}
\multirow{1}{*}{\centering MoViNet-A0 }
  & U 5x5 & 3.1M & \rev{62.17} & \rev{39.25} \\

     \noalign{\smallskip}
\multirow{1}{*}{\centering MoViNet-A0 }
  & D 16x5 & 3.1M & \rev{64.17} & \rev{40.68} \\

\bottomrule
\end{tabular}
\end{center}
\vspace{-5mm}
\end{table}
\setlength{\tabcolsep}{1.4pt}

\setlength{\tabcolsep}{7pt}
\begin{table}[t]
\begin{center}
\vspace{3mm}
\caption{Top-1 \textit{mean} accuracy ($\%$), over all $D_i \rightarrow D_j$ combinations on both seen and unseen test sets in both \textit{offline-trimmed} setting and \textit{streaming-trimmed} setting}
\label{tab:offlinevsstream}
\begin{tabular}{lccccc}
\toprule\noalign{\smallskip}
\multicolumn{5}{c}{\normalsize\textsc{EPIC-Kitchens 55}} \\
\noalign{\smallskip}
\cline{1-5}
\noalign{\smallskip}
Network & Mode & Sampling & Seen & Unseen \\

  \hline
 \noalign{\smallskip}
\multirow{1}{*}{\centering I3D  }
 & Offline  & D 16x5 & 67.08 & 42.42 \\

 \noalign{\smallskip}
\multirow{1}{*}{\centering X3D-XS  }
 & Offline  & D 16x5 & 48.45 & 32.66 \\

 \noalign{\smallskip}
\multirow{1}{*}{\centering MoViNet-A0  }
 & Offline  & D 16x5 & \rev{64.17} & \rev{40.68} \\

\hline

 \noalign{\smallskip}
\multirow{1}{*}{\centering I3D  }
 & Streaming & All Stream & \textbf{63.38} & \textbf{40.57} \\

 \noalign{\smallskip}
\multirow{1}{*}{\centering X3D-XS  }
 & Streaming & All Stream & 43.37 & 32.31 \\

 \noalign{\smallskip}
\multirow{1}{*}{\centering MoViNet-A0  }
 & Streaming & All Stream & 62.24 & 39.59 \\

\bottomrule
\end{tabular}
\end{center}
\vspace{-5.5mm}
\end{table}

\setlength{\tabcolsep}{1.4pt}

\begin{figure}[t!]
\vspace{2mm}
\centering
\hspace{-15px}
\begin{minipage}[t]{\columnwidth}
     \resizebox {\columnwidth} {!} {
        \begin{tikzpicture}
        \pgfplotsset{every axis legend/.append style={
at={(0.5,1.03)},
anchor=south}}

        \begin{axis}[
          enlargelimits=false,
          ylabel={Top-1 Accuracy (\%)},
          xlabel={Percentage (\%)},
           xmin=0, xmax=100,
           ymin=26, ymax=68,
          xtick={0,20,40,60,80,100},
          ytick={30,40,50,60},
          xmajorgrids=true,
          ymajorgrids=true,
          grid style=dashed,
          width=15cm,
          height=8cm,
          legend columns=3,
          legend style={font=\LARGE},
          label style={font=\LARGE},
          tick label style={font=\LARGE},
        legend style={draw=none},
        every axis plot/.append style={ultra thick},
        every mark/.append style={mark size=50pt}
        ]
        
        \addplot[
          color=MaterialBlue800]
        table[x index=0,y index=1,col sep=comma]
        {tex/Tables/Tab/I3D_full_sup.txt};
               \addlegendentry{I3D-(seen)}

       \addplot[
          color=MaterialOrange800]
        table[x index=0,y index=1,col sep=comma]
        {tex/Tables/Tab/movinet_sup.txt};
               \addlegendentry{MoViNet-(seen)} 
               
        \addplot[
          color=greeno]
        table[x index=0,y index=1,col sep=comma]
        {tex/Tables/Tab/X3D_sup.txt};
               \addlegendentry{X3D-(seen)}


        \addplot[
          color=MaterialBlue800, dashed]
        table[x index=0,y index=1,col sep=comma]
        {tex/Tables/Tab/I3D_full_cross.txt};
       \addlegendentry{I3D-(unseen)}


        \addplot[
          color=MaterialOrange800, dashed]
        table[x index=0,y index=1,col sep=comma]
        {tex/Tables/Tab/movinet_cross.txt};
       \addlegendentry{MoViNet-(unseen)}


        \addplot[
          color=greeno, dashed]
        table[x index=0,y index=1,col sep=comma]
        {tex/Tables/Tab/X3D_cross.txt};
       \addlegendentry{X3D-(unseen)} 

        \end{axis}
        
        \end{tikzpicture}%
      }

    \end{minipage}

\caption{MoViNet, X3D and I3D performance with respect to the percentage of video observed}
\vspace{-0.4cm}
\label{fig:videoperc}
\end{figure}
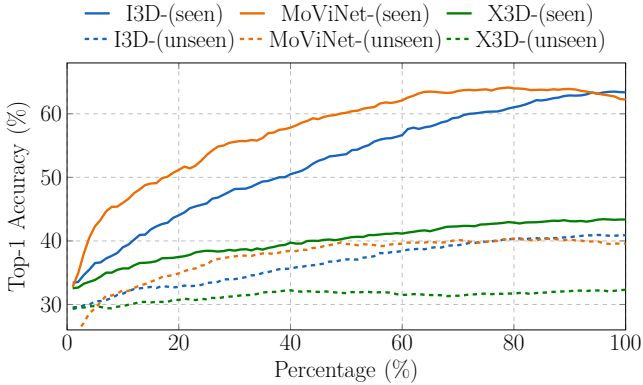

\section{Experiments}
\begin{figure}[t!]
\vspace{2mm}
\centering
\hspace{-15px}
\begin{minipage}[t]{0.235\textwidth}
     \resizebox{\textwidth}{!} {
        \begin{tikzpicture}
        \pgfplotsset{every axis legend/.append style={
at={(0.41,1.03)},
anchor=south}}
        \begin{axis}[
          enlargelimits=false,
          ylabel={Accuracy (\%)},
          xlabel={},
           xmin=4, xmax=68,
           ymin=22, ymax=50,
          xtick={0,8,16,32,48,64},
          ytick={20,25,30,35,40,45,50},
          ymajorgrids=true,
          grid style=dashed,
          width=6.7cm,
          height=6.25cm,
          legend columns=1,
        legend style={draw=none},
        every axis plot/.append style={ultra thick},
        every mark/.append style={mark size=50pt}
        ]
        \addplot[
          color=MaterialBlue800,
          mark=square*]
        table[x index=0,y index=1,col sep=comma]
        {tex/Tables/Tab/unseen-clean-static-MoViNet.txt};
        \addlegendentry{Unseen-MoViNet-(SBL)} 
        \addplot[
          color=MaterialOrange800,
          mark=oplus*]
        table[x index=0,y index=1,col sep=comma]
        {tex/Tables/Tab/seen-clean-static-MoViNet.txt};\addlegendentry{Seen-MoViNet-(SBL)} 
        \addplot[dotted,
          color=MaterialBlue800,]
        table[x index=0,y index=1,col sep=comma]
        {tex/Tables/Tab/unseen-noclean-movinet.txt};
        \addlegendentry{Unseen-MoViNet-(w/o SBL)} 
        \addplot[dotted,
          color=MaterialOrange800,]
        table[x index=0,y index=1,col sep=comma]
        {tex/Tables/Tab/seen-noclean-MoViNet.txt};
        \addlegendentry{Seen-MoViNet-(w/o SBL)} 
        \end{axis}
        \end{tikzpicture}%
      }
\end{minipage}
\begin{minipage}[t]{0.235\textwidth}
     \resizebox {\textwidth}{!} {
        \begin{tikzpicture}
        \pgfplotsset{every axis legend/.append style={
at={(0.41,1.03)},
anchor=south}}
        \begin{axis}[
          enlargelimits=false,
          ylabel={Accuracy (\%)},
          xlabel={},
           ymin=30, ymax=58,
           xmin=2.5, xmax=10.5,
          xtick={3,4,5,6,10},
          ytick={33,38,43,48,53, 58},
          ymajorgrids=true,
          grid style=dashed,
           width=6.7cm,
          height=6.25cm,
          legend columns=1,
       legend style={draw=none},
        every axis plot/.append style={ultra thick},
        every mark/.append style={mark size=50pt}
        ]
        \addplot[
          color=MaterialBlue800,
          mark=square*]
        table[x index=0,y index=1,col sep=comma]
        {tex/Tables/Tab/unseen-clean-dynamically-MoViNet.txt};\addlegendentry{Unseen-MoViNet-(DBL)} 
        \addplot[
          color=MaterialOrange800,
          mark=oplus*]
        table[x index=0,y index=1,col sep=comma]
        {tex/Tables/Tab/seen-clean-dynamically-MoViNet.txt}; \addlegendentry{Seen-MoViNet-(DBL)} 
        \addplot[dotted,
          color=MaterialBlue800,
          mark=none]
        table[x index=0,y index=1,col sep=comma]
        {tex/Tables/Tab/clean-statically-MoViNet-unseen.txt};\addlegendentry{Unseen-MoViNet-(SBL)} 
        \addplot[dotted,
          color=MaterialOrange800,
          mark=none]
        table[x index=0,y index=1,col sep=comma]
        {tex/Tables/Tab/clean-statically-MoViNet-seen.txt};\addlegendentry{Seen-MoViNet-(SBL)} 
        \end{axis}
        \end{tikzpicture}%
     }
\end{minipage}
\caption{Top-1 \textit{mean} accuracy ($\%$), over all $D_i \rightarrow D_j$ combinations on both seen and unseen  test sets in \textit{online-trimmed} setting.
Left) \textbf{Static boundary localization (SBL)} with different values for the clean buffer.
Right)
\textbf{Dynamic boundary localization (DBL)} with different threshold-values. }
\label{fig:dynamically_clean}
\label{fig:clean_static}
\end{figure}
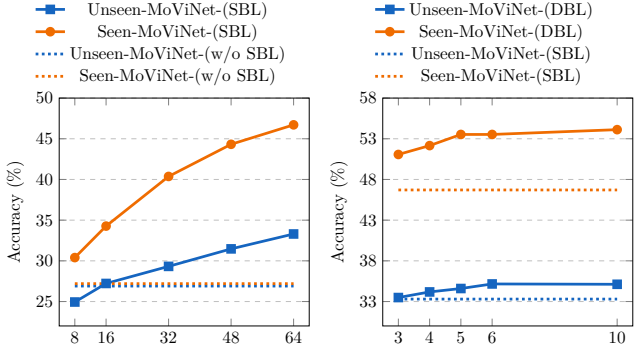

In Table \ref{tab:backbone}, we compare two families of recently  designed tiny-networks with popular architectures used in action recognition, examining various factors such as different pretrains, sampling methodologies, and amount of params. Then, in Table \ref{tab:offlinevsstream}, we analyze the performance in the streaming scenario, displaying a plot of the models' accuracy vs the portion of the video observed (Fig \ref{fig:videoperc}). Fig \ref{fig:clean_static} shows the effects of the action detection algorithms used to move from the streaming \rev{to the online inference scenario}. Finally, in Fig \ref{fig:final_results}, we test the performance with untrimmed data, demonstrating how our two-fold aggregator ($A^{2}$) method grants a more robust solution with a little increase in parameters. The Table \ref{tab:device} illustrates the impact of performance in terms of latency, power consumption, and other critical characteristics for a designed device. 

\textit{Impact of footprint on model accuracy.}
MoViNet and X3D are the two tiny architecture included in our benchmark to compare smaller models w.r.t standard action recognition networks. Interestingly, for X3D the tiny model size appears to have a negative impact on the final results, showing the lowest accuracy. It also suffers significantly from the transition from uniform to dense sampling (U$\rightarrow$D). MoViNet, on the other hand, appears to be the preferable alternative, showing more notable results in both seen and unseen settings. Noteworthy, we also observed higher robustness to the shift in sampling from uniform to dense (U$\rightarrow$D). All those considerations motivate our focus on MoViNet in this work.

\textit{The importance of seen-unseen accuracy.}
In contrast with the standard benchmark in action recognition, in our analysis we conduct experiments considering two different scenarios: seen and unseen. 
Indeed, looking only at the performances in the seen scenario, it seems that MoViNet obtains lower results compared to the TSM (\rev{62.45\%} and 71.48\% respectively).
Instead, when tested on unseen data distribution, we have a significant gain in performance of MoViNet w.r.t. TSM and I3D. 
In particular, the MoViNet results with uniform or dense strategy are quite similar (\rev{39.25\%} and \rev{40.68\%} respectively). It is also worth noticing that, with more frames, MoViNet results in unseen scenarios improve considerably (see Table \ref{tab:offlinevsstream}).

\textit{Offline $\xrightarrow[]{}$ \textit{Streaming}.}  
Table \ref{tab:offlinevsstream} shows the results in these two distinct settings. \rev{It is interesting to observe that MoViNet is the model that better exploits the continuous stream of data, obtaining the smallest deterioration in performance equal to 2\% and 1\% in seen and unseen scenarios respectively, whereas the other two networks show a much bigger decrease in performance.} This behavior is caused by the buffer implementation used in the MoViNet streaming version, which enables the simulation of a receptive field as large as the entire input video, while processing frames one-by-one ($T_s = 1$). On the contrary, I3D and X3D take as input block of 16 frames ($T_{s}$ = 16), which requires the recomputation of overlapping frames activations and may limit the total efficiency of the models. 
\textit{Streaming $\xrightarrow[]{}$ \textit{Online}.}
As discussed before, the standard action recognition protocol assumes available the knowledge of the action boundary as a prior-knowledge for the correct restart of the averaging output, to obtain video level prediction for architectures such as I3D or X3D, or to properly reset the buffer mechanism for MoViNet. In other words, ``cleaning'' the prior encoding for the new one is necessary to produce an accurate prediction for the current action. At this stage of our investigation, we assess how much the typical action recognition architectures rely on the action's boundary and how their performance is affected by the absence of this supervision knowledge. 

\textit{Dependency from the actions boundary. }
In Fig. \ref{fig:videoperc} we plot the accuracy of the models as a function of the percentage of the video observed. From this chart, we notice that the use of the last portion of the video does not provide a gain in accuracy, and after the 85\% of the video, no substantial improvement is obtained.
Similar observations can be made for the initial part of the video.
Interestingly, the performances of the tiny model X3D in the initial part of the observed video are very close to the final one, revealing a tendency to privilege appearance information with respect to motion information. Instead, the performance gap of MoViNet and I3D from the first portion of the video observed and after viewing 60\%–80\% of the data, confirms that their prediction is based more on the motion. This behavior is consistent with the more robust results in unknown conditions (unseen), where the appearance-based solution suffers more due to the fact that the appearance characteristics of the scene (texture, light condition, etc.) changes more among the environments with respect to the motion.

\textit{Effects of no supervision on actions boundaries.}
The loss of knowledge on actions boundaries requires a solution to automatically identify action changes. In this section, we discuss the performance of the strategy presented in section \ref{eventboundarydetection}, comparing the results with a static solution (Fig. \ref{fig:clean_static}). The latter is based on the ``naive'' assumption that all sample lengths are nearly equivalent, and as a result, it assumes that a new action is ``discovered'' at each $k$ frame. For both the solutions, we report a sensibility analysis on the number of frames for the static solution (SBL), and on the threshold value for the dynamic one (DBL), in both seen and unseen settings. According to Fig. \ref{fig:clean_static}, MoViNet with a low-loaded aggregator is unreliable; indeed, the results are lower than those without a clean one. Furthermore, by raising the buffer load, i.e., forwarding more frames, it increases its performance. A significant improvement of the dynamic strategy over the static one is also noticeable in Fig \ref{fig:dynamically_clean}. Moreover, MoViNet performance appears to be not sensitive to proper threshold values; indeed, the improvements of the DBL solution are always better compared to the best results of the SBL solution.
\begin{figure}[t!]
\centering
\begin{minipage}[t]{0.23\textwidth}
   \resizebox {\columnwidth} {!}{
    \centering
%
%

\begin{tikzpicture}
\centering
\begin{axis}[
symbolic x coords={\cite{du2022fast}, $A$, \textbf{$A^{2}$}, \textcolor{battleshipgrey}{S}},
	ylabel= Seen Acc. (\%),
	enlargelimits=false,
	ybar=0.5pt, enlarge x limits=0.15,
	xtick=data,
	ymin=28.0,
    ymax=66,
    ymajorgrids=true,
    legend style={draw=none},
          grid style=dashed,
          width=5.9cm,
          height=5cm,
         legend style={at={(0.5,-0.28)},
	    anchor=north,legend columns=2}, 
        every mark/.append style={mark size=7pt}
]
\addplot [fill=MaterialBlue600,]
	coordinates {
	(\cite{du2022fast},50.1)
	($A$,52.4)
	(\textbf{$A^{2}$},60.38)
	(\textcolor{battleshipgrey}{S},64.03)
	};
\addlegendentry{I3D} 

\addplot [fill=MaterialOrange600,]
	coordinates {
	(\cite{du2022fast},44.61)
	($A$,48.17 )
	(\textbf{$A^{2}$},52.87)
(\textcolor{battleshipgrey}{S},55.6)
	};
\addlegendentry{MoViNet}

\end{axis}

\end{tikzpicture}
    }
    \vspace{-0.9cm}
\end{minipage}    
\begin{minipage}[t]{0.23\textwidth}
   \resizebox {\columnwidth} {!}{
    \centering
%
%

\begin{tikzpicture}
\centering
\begin{axis}[
symbolic x coords={\cite{du2022fast}, $A$, \textbf{$A^{2}$}, \textcolor{battleshipgrey}{S}},
		ylabel= Unseen Acc. (\%),
	enlargelimits=false,
	ybar=0.5pt, enlarge x limits=0.15,
	xtick=data,
	ymin=22.0,
    ymax=42,
    ymajorgrids=true,
    legend style={draw=none},
          grid style=dashed,
            width=5.9cm,
          height=5cm,
         legend style={at={(0.5,-0.28)},
	    anchor=north,legend columns=2}, 
        every axis plot/.append 
        every mark/.append style={mark size=7pt}
]

\addplot [fill=MaterialBlue600,]
	coordinates {
	(\cite{du2022fast},36.74)
	($A$,36.12)
	(\textbf{$A^{2}$},39.28)
	(\textcolor{battleshipgrey}{S},40.57)
	};
\addlegendentry{I3D}

\addplot [fill=MaterialOrange600,]
	coordinates {
	(\cite{du2022fast},32.88)
	($A$,32.51 )
	(\textbf{$A^{2}$},35.19)
	(\textcolor{battleshipgrey}{S},37) };
\addlegendentry{MoViNet} 

\end{axis}
\end{tikzpicture}
    }
    \vspace{-1.4cm}
\end{minipage}    
\caption{\rev{We present the results in an \textit{online-untrimmed} scenario using three different approaches: ABD \cite{du2022fast} as a secondary stream to identify the boundaries, our DBL technique with a single aggregator ($A$), and our DBL technique with a two-fold aggregator (\textbf{$A^{2}$}). We also report the results obtained in a streaming scenario (\textcolor{battleshipgrey}{S}) as an upper bound reference.}}
\label{fig:final_results}
\end{figure}
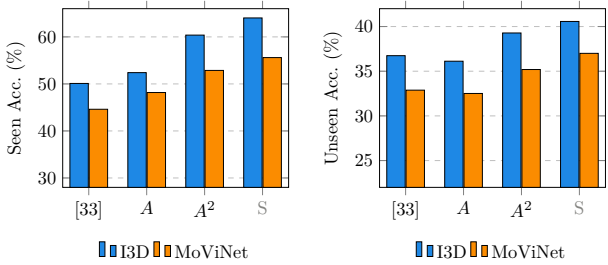
    
\begin{figure}[t!]
\centering
\hspace{-15px}
\begin{minipage}[t]{0.23\textwidth}
     \resizebox{\textwidth}{!} {
        \begin{tikzpicture}
        \pgfplotsset{every axis legend/.append style={
at={(0.41,1.03)},
anchor=south}}
        \begin{axis}[
          enlargelimits=false,
          ylabel={Accuracy (\%)},
          xlabel={$\delta$},
          xmin=-5, xmax=55,
          ymin=49, ymax=63,
          xtick={1,5,10,20,30,40,50},
          ytick={50, 52, 54,56, 58, 60, 62},
          ymajorgrids=true,
          grid style=dashed,
          height=5.25cm,
          legend columns=1,
        legend style={draw=none},
        every axis plot/.append style={ultra thick},
        every mark/.append style={mark size=50pt}
        ]
        \addplot[
          color=MaterialBlue800,
          mark=square*]
        table[x index=0,y index=1,col sep=comma]
        {data/i3d_sup_delta.csv};
        \addlegendentry{Seen-I3D} 
        \addplot[
          color=MaterialOrange800,
          mark=oplus*]
        table[x index=0,y index=1,col sep=comma]
        {data/mov_sup_delta.csv};\addlegendentry{Seen-MoViNet} 
        \end{axis}
        \end{tikzpicture}%
      }
\end{minipage}
\begin{minipage}[t]{0.23\textwidth}
     \resizebox {\textwidth}{!} {
        \begin{tikzpicture}
        \pgfplotsset{every axis legend/.append style={
at={(0.41,1.03)},
anchor=south}}
        \begin{axis}[
          enlargelimits=false,
          ylabel={Accuracy (\%)},
          xlabel={$\delta$},
          xmin=-5, xmax=55,
          ymin=33.5, ymax=40.5,
          xtick={1,5,10,20,30,40,50},
          ytick={34,35,36,37, 38, 39,40},
          ymajorgrids=true,
          grid style=dashed,
          height=5.25cm,
          legend columns=1,
        legend style={draw=none},
        every axis plot/.append style={ultra thick},
        every mark/.append style={mark size=50pt}
        ]
        \addplot[
          color=MaterialBlue800,
          mark=square*]
        table[x index=0,y index=1,col sep=comma]
        {data/i3d_cross_delta.csv};\addlegendentry{Unseen-I3D} 
        \addplot[
          color=MaterialOrange800,
          mark=oplus*]
        table[x index=0,y index=1,col sep=comma]
        {data/mov_cross_delta.csv}; \addlegendentry{Unseen-MoViNet} 
        \end{axis}
        \end{tikzpicture}%
     }
\end{minipage}
\caption{\rev{Analysis of the effects of the delay parameter $\delta$. We report Top-1 \textit{mean} accuracy ($\%$) in \textit{online-untrimmed} setting.}}
\label{fig:delta_ablation}
\vspace{-4.2mm}
\end{figure}
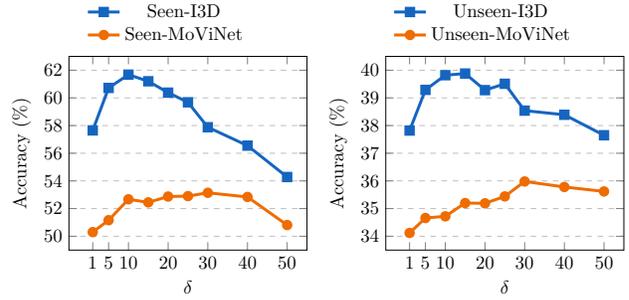
\textit{Trimmed $\xrightarrow[]{}$ \textit{Untrimmed}.}
In Fig \ref{fig:final_results} we show the results in an untrimmed online scenario. We compare the performance of our DBL approach with  single ($A$) and two-fold  ($A^{2}$) aggregator, to the recently proposed technique ABD \cite{du2022fast}, exploiting it as a secondary stream to identify the boundary and provide the action boundary to the primary model of classification. \rev{For ABD, we used the original online implementation, with both NMS and filter windows size equal to 50.} Furthermore, we report, as a reference, untrimmed streaming (S) results, i.e., experiments in which the real boundary of the action is used as prior knowledge. 
We present the performance of the DBL technique and two-fold aggregator using I3D to demonstrate that the proposed approach is scalable and model agnostic. Indeed the improvement of $A^{2}$ is remarkable and the results obtained for both the architecture are comparable with the streaming scenario. Moreover, the solution with a single aggregator performs \rev{similarly} to the competitor ABD, without using a secondary stream for the boundary localization. 
Finally, the improvements of our solution $A^{2}$ with respect to the ABD are consistent across scenarios and models. 
\rev{To provide a comprehensive analysis, we conducted an ablation study on the delay hyperparameter $\delta$. The results are presented in Fig. \ref{fig:delta_ablation} and confirm that estimating $\delta$ as the average overlap of actions at the desired frame rate is a reliable approach.}

\setlength{\tabcolsep}{2.0pt}

\begin{table}[t]
\vspace{3mm}
\begin{center}

\caption{MACs, FPS (Hz), Latency (ms, \rev{inference time}) and Energy(watt) on different devices.}
\label{tab:device}
\begin{tabular}{lccccc}
\toprule\noalign{\smallskip}
\multicolumn{6}{c}{\normalsize\textsc{On Device}} \\
\noalign{\smallskip}
\cline{1-6}
\noalign{\smallskip} 
Network & Device & MACs & FPS & Lat.(ms) & Power(watt) \\
\hline
\noalign{\smallskip}
\multirow{1}{*}{\centering I3D }
 &  2080 Ti  & \rev{$270e^8$} & 110 & $9.1$ & 53.7  \\
\multirow{1}{*}{\centering MoViNet }
 &  2080 Ti  & $0.47e^8$ & 781 & $1.3$  &  52 \\
 \hline
\noalign{\smallskip}
\multirow{1}{*}{\centering I3D }
 &   MX350 & \rev{$270e^8$} & \textcolor{red}{\textbf{15}} & $65.7$ & 24.9  \\
\multirow{1}{*}{\centering MoViNet }
 &   MX350 & $0.47e^8$ & 169 & $5.9$  & 11.5 \\
 \hline
\noalign{\smallskip}
\multirow{1}{*}{\centering I3D } 
&   Jetson Nano  & \rev{$270e^8$} & \textcolor{red}{\textbf{3}} & $393.7$ & 3.7 \\
\multirow{1}{*}{\centering MoViNet }
 &  Jetson Nano  & $0.47e^8$ & \textcolor[rgb]{0,0.502,0}{\textbf{56}} &  $17.9$ & \textcolor[rgb]{0,0.502,0}{\textbf{2.4}}  \\
\bottomrule
\end{tabular}
\end{center}
\vspace{-7mm}
\end{table}
\setlength{\tabcolsep}{1.1pt}

\textit{Edge Deployment.}
In Table \ref{tab:device} we show MACs, FPS Latency and Energy on different devices. These metrics are obtained from models deployed on each different hardware through the usage of TensorRT. Power is measured with a power meter, subtracting the static power. This analysis focuses on how hardware constraints affect the applicability of the existing model for action recognition on real device. Indeed, when the I3D model moved from a high-performance GPU (2080 Ti) to a laptop GPU (MX350) and, to an edge device (NVIDIA Jetson Nano), it used more energy, falling short of the required FPS threshold for identifying human motion (up to 20-30 FPS \cite{song2016fast}). Instead, in the case of MoViNet, the minimal number of model parameters ensures appropriate FPS (twice as needed), allowing the use of  two-fold aggregator technique in online \rev{inference} scenario. 

\section{Conclusions}
\rev{The purpose of this work is to investigate and highlight the limitations that mainstream egocentric vision models show in realistic usecases, where computational time and power are limited. We promote a new line of research for FPAR, which considers real-world application limits such as hardware restrictions, cross-domain scenarios, and online inference on untrimmed data.\\
We provide: i) a new benchmark to assess the challenges of real world deployment, and ii) a novel approach capable to bring FPAR models on low-power devices (edge computing), tackling the presence of overlapping actions and the absence of supervision on action boundaries for real world usage. 
In light of the challenges discussed in this work, we encourage future researchers to devote attention to designing innovative approaches that allow real-time adaptation of the model on the edge during the processing of untrimmed videos, particularly in the presence of changes in environmental conditions.}


\bibliographystyle{ieeetr}
\bibliography{IEEEabrv, egbib.bib}

\end{document}